\documentclass[letterpaper, 10 pt, conference]{ieeeconf} % use this command before submit
\IEEEoverridecommandlockouts                              % This command is only needed if 
\usepackage{multirow}
\usepackage{algorithm}
\usepackage{cite}	
\usepackage{algorithmic}
\usepackage{graphicx}
\usepackage{times}
\usepackage{amsmath}
\usepackage{amssymb}
\usepackage{subfigure}
\usepackage{graphics} % for pdf, bitmapped graphics files
\usepackage[utf8]{inputenc}
\usepackage{boldline}
\usepackage{multicol}
\usepackage{lipsum}
\usepackage[pdfstartview=FitH,bookmarksnumbered=true,colorlinks,bookmarksopen=true]{hyperref}
\newcommand{\etal}{\textit{et al}.~}
\newcommand{\ie}{\textit{i}.\textit{e}.~}

\title{\LARGE \bf
Making Sense of Audio Vibration
\\ for Liquid Height Estimation in Robotic Pouring}
\author{Hongzhuo Liang$^{1}$, Shuang Li$^{1}$, Xiaojian Ma$^{2,3}$, \\
Norman Hendrich$^{1}$, Timo Gerkmann$^{4}$, Fuchun Sun$^{3}$, Jianwei Zhang$^{1}$
\thanks{$^{1}$TAMS (Technical Aspects of Multimodal Systems), Department of Informatics, Universit\"{a}t Hamburg}
\thanks{$^{2}$Center for Vision, Cognition, Learning, and Autonomy, Department of Statistics, University of California, Los Angeles}
\thanks{$^{3}$Beijing National Research Center for Information Science and Technology (BNRist), State Key Lab on Intelligent Technology and Systems, Department of Computer Science and Technology, Tsinghua University}
\thanks{$^{4}$SP (Signal Processing), Department of Informatics, Universit\"{a}t Hamburg}
}
\begin{document}
\maketitle
\thispagestyle{empty}
\pagestyle{empty}

%%%%%%%%%%%%%%%%%%%%%%%%%%%%%%%%%%%%%%%%%%%%%%%%%%%%%%%%%%%%%%%%%%%%%%%%%%%%%%%%
\begin{abstract}
In this paper, we focus on the challenging perception problem in robotic pouring. Most of the existing approaches either leverage visual or haptic information. However, these techniques may suffer from poor generalization performances on opaque containers or concerning measuring precision. To tackle these drawbacks, we propose to make use of audio vibration sensing and design a deep neural network PouringNet to predict the liquid height from the audio fragment during the robotic pouring task. PouringNet is trained on our collected real-world pouring dataset with multimodal sensing data, which contains more than 3000 recordings of audio, force feedback, video and trajectory data of the human hand that performs the pouring task. Each record represents a complete pouring procedure. We conduct several evaluations on PouringNet with our dataset and robotic hardware. The results demonstrate that our PouringNet generalizes well across different liquid containers, positions of the audio receiver, initial liquid heights and types of liquid, and facilitates a more robust and accurate audio-based perception for robotic pouring.
% In this paper, we present an audio-based scheme PouringNet for detecting the liquid height in robotic pouring tasks.
% PouringNet handles the correlations in audio sequences and the liquid height in real-time, using spectrograms as input and the height of the air column in the cups as output. 
% PouringNet is trained on a multimodal dataset, which contains 3000 records of audio data, force feedback, trajectories during human pours.
% The network evaluation results verify the eligibility of PouringNet, while our robotic experiments on pouring tasks with unseen target containers, varied initial heights, different liquids, manifold position of audio sources show PouringNet is scalable while being robust to the noisy background.
\end{abstract}

\section{Introduction}
% what is robotic pouring, what is the perception in robotic pouring? why it is challenging?
% what it current method? what are their drawbacks?
Robotic pouring~\cite{tamosiunaite2011learning} is a crucial robotic task in both domestic and industrial environments. 
In a nutshell, a robot is required to pour a liquid from one container to another while preventing it from spilling. Therefore, the robust and accurate perception will play an essential role in this task, especially in estimating the liquid height in the target container. Recent approaches to solving this perception problem mostly rely on visual sensing~\cite{visualpouring1,visualpouring2,visualpouring3,visualpouring4}. 
By leveraging a camera situated in front of the target container, the current liquid height can be regressed from the visual features of the captured image. 
However, these approaches cannot generalize to opaque containers since the liquid height cannot be seen or could suffer from poor estimation error. On the other hand, haptic sensing is another important modality for the perception of robotic pouring.
For example, when the force and torque feedback on the manipulator is available, we can either estimate the volume of liquid being poured or directly learn a pouring policy in an end-to-end manner~\cite{rozo2013force}.
However, the correlation between haptic information and the pouring liquid can be rather complicated and are varied among different end effectors and containers.
%This limits the generalizing performance of the policies.
These drawbacks in existing perception methods suggest that the robust and accurate perception in robotic pouring still remains an open problem.
To sum up, the major challenges in the perception for robotic pouring are twofold: 1. \textbf{Generalization}, the perception in robotic pouring should be able to generalize to different containers, liquid type, and liquid status. 2. \textbf{Precision}, the estimation result, \ie, the prediction of liquid height, should be accurate enough to satisfy the requirement in pouring task.
%and is \textbf{not} an easy task by any measure.

%Robotic pouring \cite{tamosiunaite2011learning} is ubiquitous in both domestic and industrial environments. Vision sensing, force sensing, and acoustic sensing all play essential roles in pouring activities \cite{pieropan2014audio, portnoy2015effect}. However, vision data is inferior to other modalities when it comes to steaming condition, pouring water to a transparent glass or a deep and opaque cup.

\begin{figure}[t]
    \centering
    \includegraphics[width=0.48\textwidth]{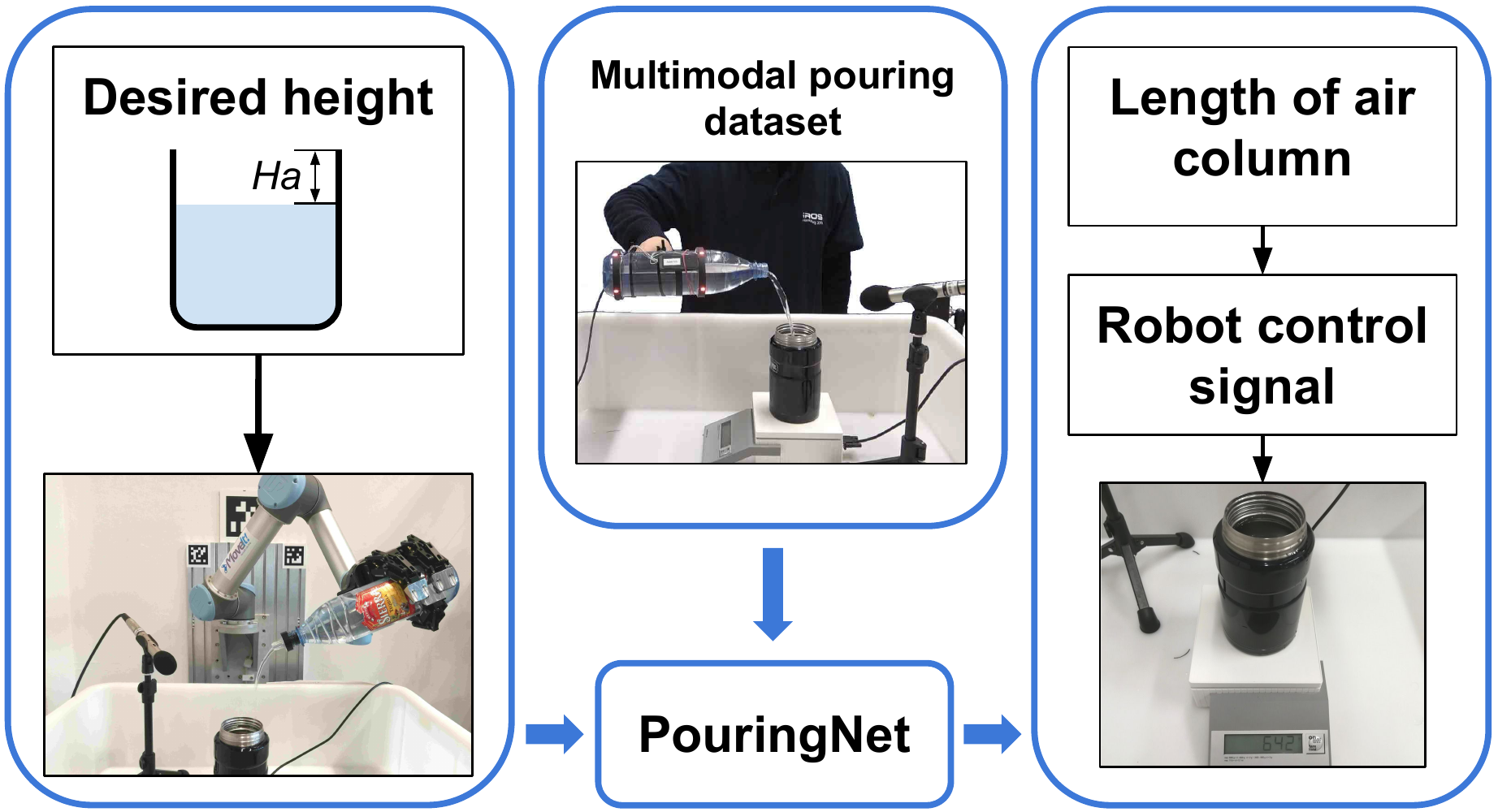}
    \caption{Our robotic pouring system. 
		(Left) Given the robot a target liquid level, audio vibrations during the pouring manipulation by a robot are recorded by a microphone then fed into PouringNet. 
		(Center) PouringNet is trained offline to predict the length of the air column of the target container from our multimodal pouring dataset.
		(Right) The length of the air column $H_a$ predicted by PouringNet is used to guide the robot's pouring control signal.}
    \vskip -0.15in
    \label{fig:overall}
\end{figure}

% why use audio? why audio is good? Any counterparts using audio? what are their drawbacks? 

To this end, we propose to tackle these issues by leveraging the modality of acoustics. Inspired by how human judge the liquid height during pouring with their hearing, we try to design a model that can estimate the position of liquid height with audio vibration. This is based on the observation that the vibrational frequency of the air in the container will change as the level of liquid rises during the pouring procedure.
Moreover, estimating liquid height using audio vibration is immediate. Thus there is no need to explicitly perform an integration, which further reduces the prediction bias and achieves more accurate results.

%Basically, sounds are not easily affected by the properties of the cup and operating environment. Studies have shown that auditory signals \cite{velasco2013sound} contain rich features in pouring task. When pouring liquid into a container, as the liquid gets more and more, two distinctive resonance frequencies will change. First, the resonance of the container itself becomes slightly lower because of the resistance to water pressure \cite{french1983vino}. Second, when the length of the air column gets shorter, the air vibrates faster, that means the resonance frequency of the air increases. Therefore, we believe that using audio sequence makes sense to estimate liquid height for different cups. Meanwhile, force and torque feedback \cite{rozo2013force} from the pouring container explicitly represents the volume of the pour-out liquid. Inspired by human experiences, we investigate to take sound modality as input for the manipulation of pouring liquid then to yield the specified height of the target container by training a recurrent network \cite{graves2013speech}.

% briefly introduce the design of PouringNet, dataset and exps (repeat the abstract)
In this paper, we introduce a deep network called PouringNet that utilizes the audio vibration to estimate liquid height for robotic pouring. Specifically, the main part of PouringNet is a recurrent neural network that maps an audio fragment into the prediction of current height of liquid height. Fig.~\ref{fig:overall} illustrates the perception pipeline in our proposed method for robotic pouring. 
There are two considerations on choosing a recurrent structure for our PouringNet. Firstly, using RNNs was already shown to be effective in audio analysis due to the inherent sequential properties of audio itself. Secondly, a recurrent structure can implicitly integrate the prior knowledge that the liquid level rises up monotonically during the pouring, which may help the model to reduce the noise and to predict more smooth results. These two considerations are verified in both our dataset and robotic experiments.

Recent successes in deep neural network based robotic perception methods~\cite{alexnet,googlenet,vgg,resnet,densenet} emphasize the importance of training on large-scale datasets. To further improve the performances of the proposed method, we built up a dataset which contains more than 3000 records of audio-frequency recordings, force and torque feedback, video flows, and motion trajectories from human demonstrations. 
%with multimodal information of %more than
%3000 records of real-world pouring. 

%In this paper, we focus on exploring an efficient learning model which could generalize the perception ability to different kinds of liquids and containers, varying initial volume and so on. To achieve this supervised learning, we aim to collect a multimodal pouring dataset and find out an indicative groundtruth. Since collecting dataset by a robot suffers the high cost, we instead collect multimodal pouring dataset by humans and to transfer the learned knowledge to robot scenario. The groundtruth not only directly correlate with the sound frequency but also is compatible for different cups. The overview of our approach is illustrated in Fig. \ref{fig:overall}. 

% contributions
In summary, our key contributions are twofold:
\begin{itemize}
\item We propose to tackle the challenging liquid height estimation task in the perception of robotic pouring by leveraging the audio vibration data and a neural network model with a recurrent structure.
%Compared to other counterparts that are either based on visual or haptic sensing~\cite{visualpouring1,rozo2013force}, our method can generalize better to different types of containers, manipulator and liquid status while the precision of predictions can be guaranteed. 
Extended experiments on dataset and robotic hardware demonstrate that our method can facilitate a robust and accurate audio-based perception for robotic pouring.
\item We built a large-scale multimodal dataset
%\footnote{The multimodal robotic pouring dataset will be open-sourced along with the PouringNet model.}
with a focus on the perception task for robotic pouring, which contains more than 3000 human pouring sequences. To the best of our knowledge, this is the first multimodal dataset for the perception task in robotic pouring.
%In addition to the considered audio data and liquid height label, data from visual and haptic sensing is also included. Although we only consider the modality of audio vibration in this paper, future work on perception methods with multimodal fusion can be facilitated with this dataset.
\end{itemize}

% Our primary contributions are: 
% %1) To our knowledge, this is the first time to use the audio vibration as input in robotic pouring. 
% 1) We propose an audio-based framework PouringNet in recurrent architecture, which learns the real-time height of the air column in pouring task.
% %We further investigate the proper architecture of PouringNet through the evaluation of height error and comparison to other baselines. 
% 2) We build a multi-modal pouring dataset that includes 3000 sequences of acoustic data, force feedback, pouring trajectories as well as the corresponding liquid height for three different cups by human.
% 3) A UR5 robot arm conducts various evaluations of generalization to pouring tasks.

\section{Related Work}
\noindent
\textbf{Robotic pouring.}
Robotic pouring has usually been implemented through generating motion trajectories or estimating specified features of the liquids or the containers as the guidance for pouring tasks.
%On the one hand, motion trajectories of the robot mainly include angular velocity, the relative displacement of the robot without any liquids spilling out.
Brandi \etal \cite{brandi2014generalizing} suggested learning pouring tasks using kinaesthetic teaching and then generalized pouring actions by computing warped parameters.
%used for matching the point cloud of the known container to the new container.
Learning dynamic pouring tasks from human demonstration was implemented in \cite{yamaguchi2015pouring, langsfeld2014incorporating}.
Since transferring compliant manipulation skills from humans to robots is always difficult, Pan \etal solved the online trajectories of the source container in simulation using a receding-horizon optimization method to handle the fluid dynamics \cite{pan2016motion}.
More recently, Do \etal \cite{dochau2018learning} solved this problem by learning a pouring policy using deep deterministic policy gradients in simulation and transferring the learned policy from simulation to a real robot.
On the other hand, some researchers focus on using different modalities to detect viscosity \cite{elbrechter2015discriminating}, height \cite{dochau2016}, the volume of the liquid or granular material \cite{audio_corl}, etc., thus relying on the perception results to perform pouring by a simple controller on the real robot. 

\noindent
\textbf{Visual sensing for robotic pouring.}
Vision is one of the commonly used modalities while pouring and in everyday life, humans also highly rely on vision when pouring water.
Obviously, vision-based perception highly depends on the lighting conditions, the color of the liquids and the shape of the target containers.
%Pithadiya \etal \cite{pithadiya2011selecting} summarily compared several edge detection techniques for the filling height inspection of target containers.
Do \etal 
advocated a probabilistic approach to estimate the liquid height based on an RGB-D camera \cite{dochau2016}. They further switched the analytical estimation approaches depending on the type of liquid and utilized a Kalman filter dealing with the uncertainties of the vision data \cite{dochau2019}. However, the mean height errors for ten pours of 3 transparent liquids were larger than 4\,mm.
Instead of directly predicting the absolute height of liquid, another popular method estimates the input volume of the liquid by analyzing the visual information of the water flow~\cite{visualpouring1}. Schenck \etal \cite{visualpouring1} used a thermal camera to generate pixel-level groundtruth data of heated water using thermal imagery. The estimation result was used to determine the water volume using both a model-based method and a neural network method.
However, this method suffers from poor estimation error due to the varied liquid types and the complex shapes of water flow. 

\noindent
\textbf{Haptic sensing for robotic pouring.}
Besides, haptic sensing, especially force and torque sensing, are also popular in the perception of robotic pouring. Specifically, force data is exerted to generate pouring trajectories by predicting the angular velocity of the pouring container in simulation \cite{huang2017learning}.
Rozo \etal \cite{rozo2013force} used a parametric hidden Markov model to retrieve joint-level commands given the force-torque inputs from the human demonstration.
Hannes \etal \cite{saal2010active} examined the viscosity estimation of the various liquids from tactile sensory data.
Although force from the pouring container could explicitly represent the volume of the pour-out liquid, force cannot measure the liquid height in an unseen target container.

\noindent
\textbf{Audio sensing for robotic pouring.}
The auditory information embodies sustainable clues when liquid or granular materials interact with other objects, such as resonance frequency and vibration behavior of the liquids and the air.
Griffith \etal \cite{griffith2012object} indicated that auditory and proprioceptive data enhance the classification tasks for the interaction between objects and water.
Sakiko \etal \cite{ikeno2015change} verified that the vibration when pouring liquid out as an audio-haptic rendition significantly affects the amount of liquid poured.
Clarke \etal \cite{audio_corl} used audio-frequency vibration generated by shaking the granular material to evaluate the weight that poured out.
However, the weight of the poured out granular materials does not allow to estimate the filling height of the target container.
Granular materials and liquids also have entirely different properties.
None of the above work explores the fact that exploiting audio vibration of the air in the target container to solves a height regression issue in robotic pouring.
%TODO error 
%Several studies have shown that in pouring tasks, multi-modal sources represent the environmental features better than one raw sensor data.
%Wu \etal \cite{wu2018liquid} presented a hierarchical long short-term memory \cite{hochreiter1997long} model which could detect if one pouring sequence is successful or failed based on a pouring dataset including visual sequences and IMU data. Unfortunately, neither their method nor the multimodel dataset involves in the robotic application at all.

In this work, we follow the perception-based methods to estimate the liquid filling level of the target containers. We take advantage of the phenomenon that the resonance frequencies implicitly relate to the length of the air column of the container and aim to design a robust model which could generalize to different situations.

\section{Data Preparation}
\subsection{Multi-modal Pouring Dataset}
Training a regression model which learns the correlations between the audio vibration and the liquid height in the receivers relies on a meaningful pouring dataset.
We collected a multi-sensor dataset by humans in a quiet environment. 
Besides the audio data and liquid height labels (measured indirectly via a digital scale under the target container), we also recorded videos of the whole pouring procedure, the trajectories of the human hand that performed the pouring by a motion tracking system, and the haptic feedback of the end effector of the manipulator via a torque sensor on the source container.
The cost of collecting dataset by humans is comparatively low compared to a robot. However, this approach brings the challenge of transferring a model learned from this dataset to robotic pouring due to the different pouring trajectories and loud background noise.

\begin{figure}[t]
	\centering
	\includegraphics[width=0.45\textwidth]{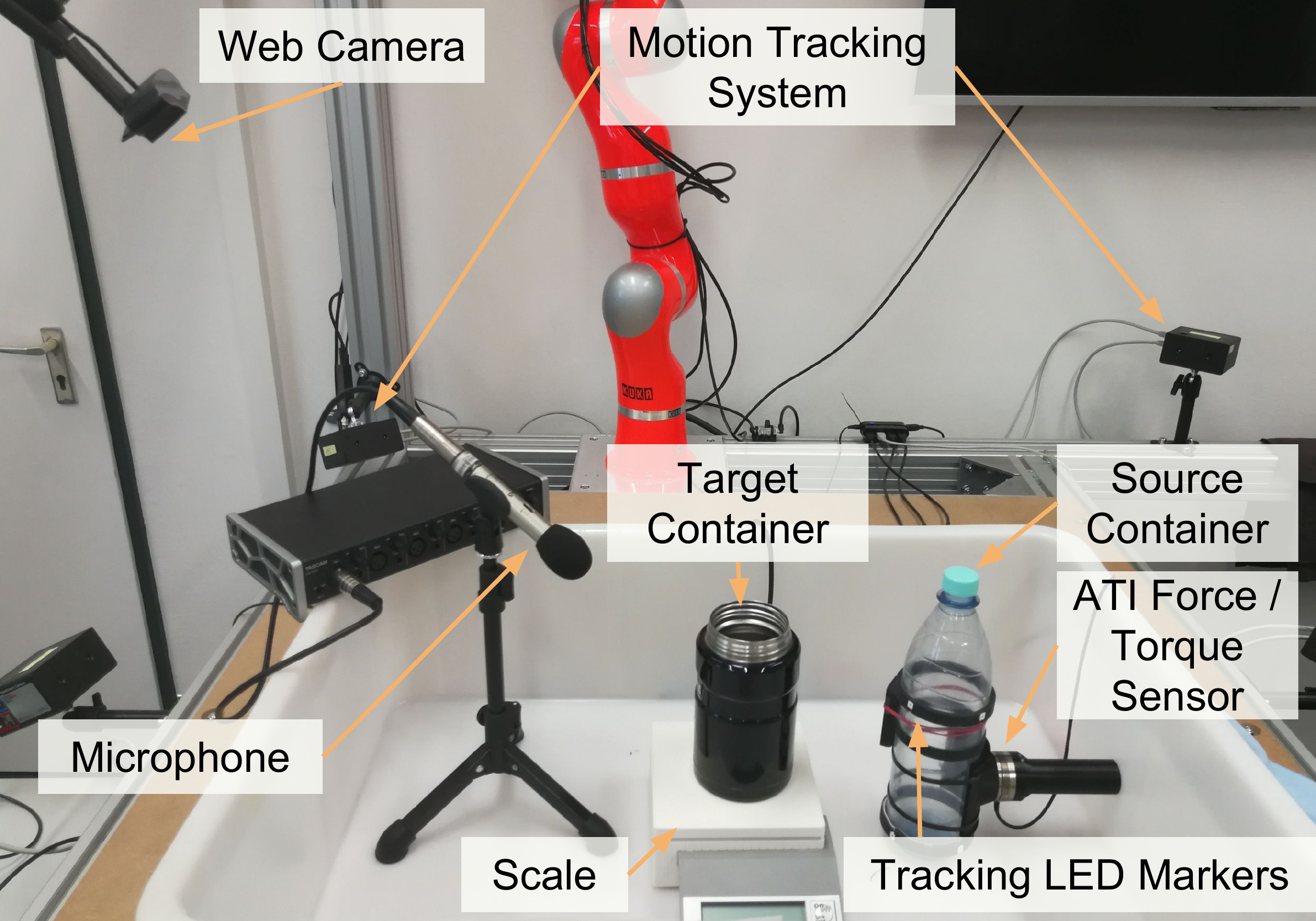}
	\caption{Pouring setup used to collect our multi-modal dataset.}
	\vskip -0.15in
	\label{fig:dataset_setup}
\end{figure}

Our dataset setup as shown in Fig. \ref{fig:dataset_setup}. It includes a source container, three different target containers (referred to as glass, thermos, and mug, shown in the first three items of Fig \ref{fig:newcups}), a Behringer B-5 microphone (44.1\,kHz), an ATI Mini40 Force / Torque sensor (500\,Hz), a Maul Logic digital scale (1\,Hz), a Logitech web camera (30\,Hz), and a PhaseSpace Impulse X2E motion tracking system (240\,Hz).
The height of the glass, thermos, and mug respectively are 127\,mm, 150\,mm and 99\,mm.
We placed the source containers relative to the bottom center of the microphone at a horizontal distance of 250\,mm and a vertical distance of 750\,mm.
For each pouring trial, the subject held the handle of the source container and started pouring task at an angle varying $8^{\circ} \pm 15^{\circ}$ and at a random position which is relative to the mouth of the target container ranging from 450\,mm to 500\,mm. Pouring during the training only involved water.

In this manner, we respectively collected 1000 trials for three target containers involving two subjects. For each trial, we took multi-modal data right before the scale reading starts to change, and right after the scale reading becomes stable. The lengths of one pouring recording varied from 4 seconds to 11 seconds.
Although we only consider the modality of audio vibration in this paper, future work on perception methods with multimodal fusion can be facilitated with our dataset.

\subsection{Data Analysis}

\noindent
\textbf{Audio-frequency data.}
When pouring the liquid into a container, as the liquid height rises, two distinctive resonance frequencies will change. First, like in an organ pipe, when the length of the air column of the container gets shorter, the air vibrates faster and the resonance frequency of the air increases.
Second, the resonance frequency of the container itself becomes slightly lower because of the resistance to water pressure \cite{french1983vino}. 

\begin{figure}[t]    
\centering
	\includegraphics[width=0.47\textwidth]{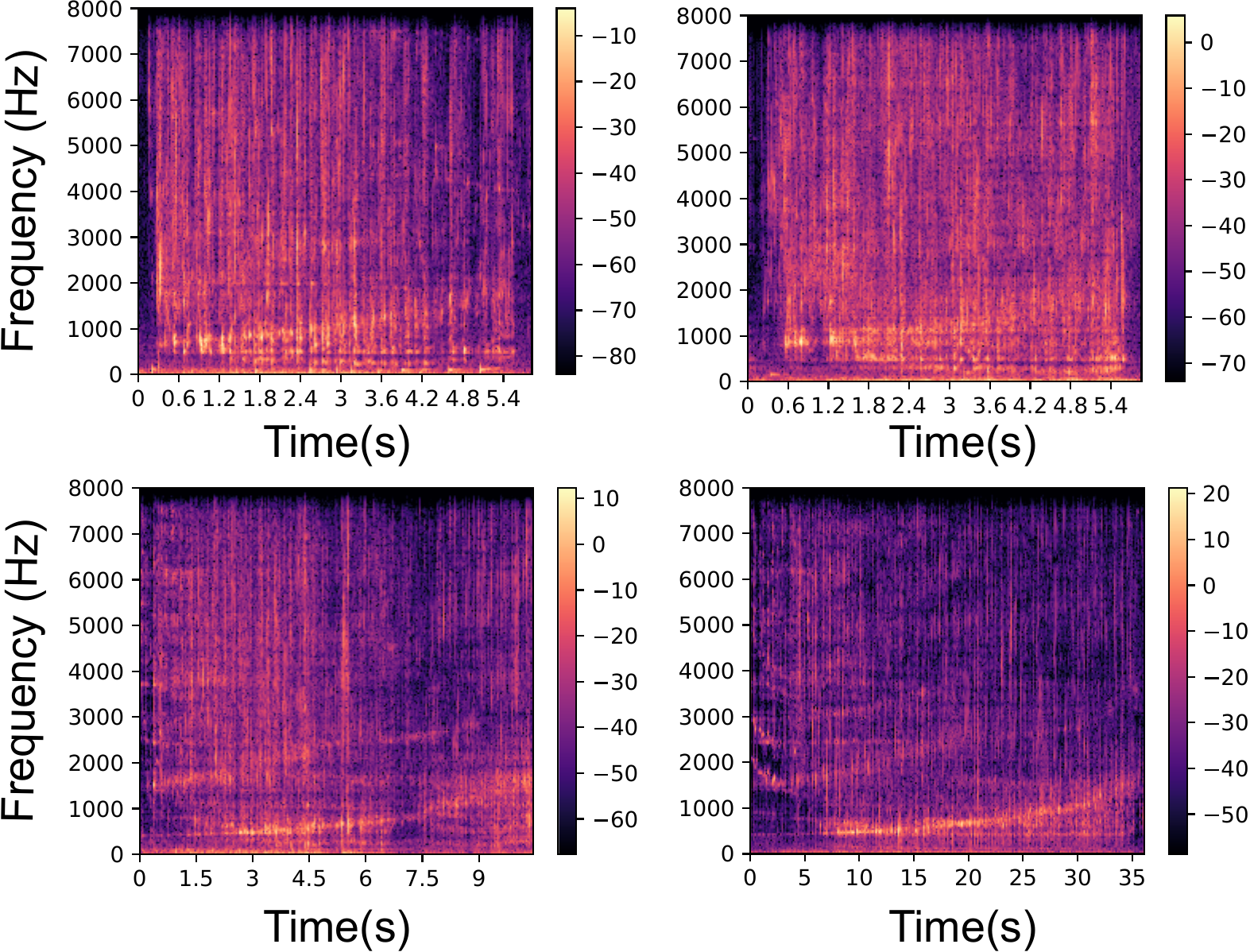}
    \caption{Examples of audio spectrograms in our dataset. (Top) From left to right, the spectrograms from pouring into the glass, and the mug. (Bottom) From left to right, the spectrograms from pouring into the thermos by source container with/without a spout in Fig. \ref{fig:spout}.}
     \vskip -0.15in
    \label{fig:audio}
\end{figure}

We resample all audio data to 16\,kHz and computed spectrograms with a window length of 0.032 seconds and a half-window overlap. We use an FFT frequency resolution of 512, and this generates audio slices with 257 descriptors. 
As four spectrogram examples shown in Fig. \ref{fig:audio}, one high-energy and rising curve between 256\,Hz-2048\,Hz is clearly visible in all spectrograms. This curve represents the resonance frequency of the air.
Meanwhile, the bottom two spectrograms in Fig. \ref{fig:audio} demonstrate that different pouring speed does not change the main theory of the resonance frequency of the air.
We propose that the rising of the resonance frequency of the air actually depends on the liquid height and not on the much heavier and more heavily damped liquid height.
Therefore, using audio vibration to estimate the real-time liquid height could make sense.
But the resonance frequency of the target container which should principally have a slight downtrend cannot be clearly seen in our dataset. 

\noindent
\textbf{Scale data.}
The weight data measured by the scale is used to calculate the liquid height.
To accomplish the real-time perception task, we deploy linear interpolation to the initial weight recordings because of the low sample rate of the scale.
We respectively generate a quadratic polynomial with fifteen pairwise weight and height measurements for each target container.
Afterward, we calculate the corresponding liquid height using the polynomial from the interpolated weights.
It follows that the height data is a $257 \times 1$ times continuous series.

\section{Pouring Network}\label{sec:net}
Our goal is to acquire the desired filling height of the liquid through learning a robust model based on audio vibration.
Although the correlations between the resonance frequency of the air and the liquid height are time-independent, the pouring task is a sequential and variable-length problem.
Instead of choosing a simple feed forward network to reach our goal, we make use of the recurrent neural network to exhibit and learn the real-time height of the liquid.
There are two popular recurrent units we can use: long short-term memory (LSTM) unit and gated recurrent unit (GRU) \cite{cho2014properties}, both of which alleviate the vanishing gradient problem in a gating mechanism.

\begin{figure}[t]
\centering
  \includegraphics[width=0.2\textwidth]{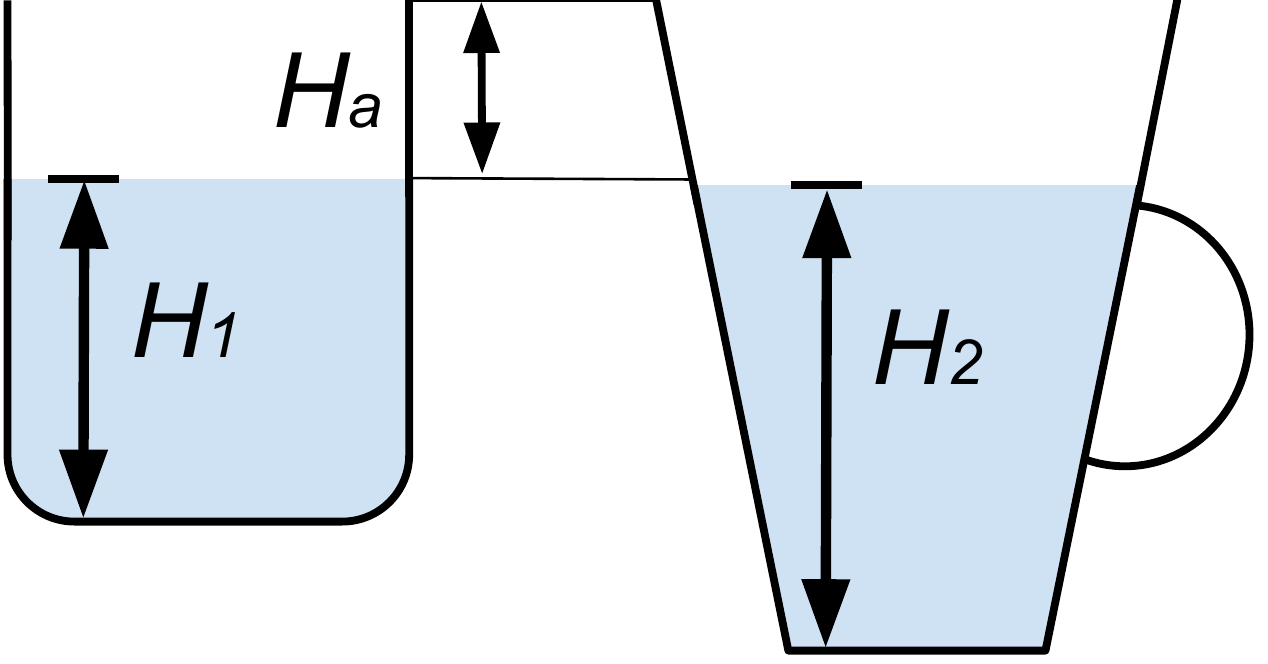}
  \caption{Different target containers are filled to different height of the liquid, but have the same length of the air column $H_a$ and a similar resonance frequency of the air when the liquid is poured into them.}
  \label{fig:cavity}
  	\vskip -0.15in
\end{figure}

Furthermore, determining a well-suited groundtruth is also crucial in supervised learning.
The physical model suggests that the increase of resonance frequency in pouring motions results from the reduced length of the air column. And this conclusion exists on different types of target containers.
Intuitively, we deduce that target containers with an air of the same height have a similar resonance frequency although they are of different shapes and different filling heights. That is to say, in our task, to make sense of audio vibration, using the length of the air column as groundtruth could be more generative and indicative than using the height of the liquid level.
For example, in Fig. \ref{fig:cavity}, two target containers are filled at different liquid heights but have an air column of the same length. Otherwise, if we take the height of liquid level as the groundtruth, we would have one label corresponding to different audio-frequency features.

\begin{figure}[t]
    \centering
    \includegraphics[width=0.46\textwidth]{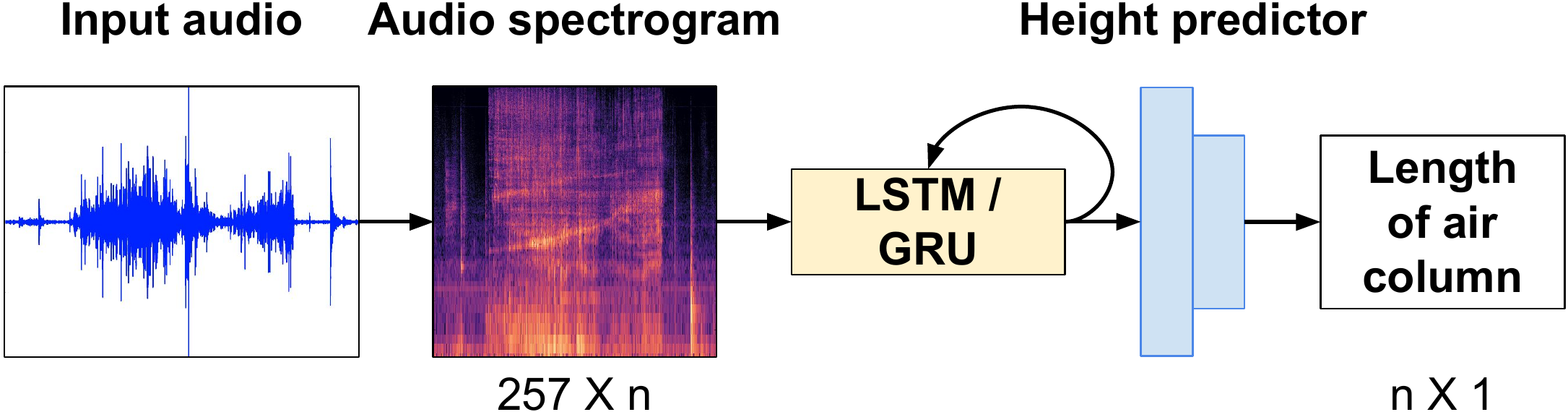}
    \caption{PouringNet architecture. The raw audio data is transformed into a spectrogram with 257 descriptors. Then the encoder module (a recurrent neural unit) is progressively fed each time slice of audio-frequency spectrogram. Finally, the height predictor module produces the 1D length of the air column of the target containers.
    The blue rectangular denotes a fully-connected layer following with a batch normalization layer and a rectified linear unit.}
    \vskip -0.15in
    \label{fig:net}
\end{figure}

With the above considerations, we design a recurrent deep network (PouringNet) $P_\theta$ to predict the length of the air column $H_a$. $\theta$ defines the parameters of our proposed PouringNet. The network architecture is shown in Fig. \ref{fig:net}.

%In order to enable our model to learn the length of the air column $H_a$ from various initial liquid heights, 
In order to augment our dataset, we randomly choose audio clips with a length of 4 seconds from one complete pouring audio sequence. The number of audio clips is proportional to the length of a pouring trial. 
We take the raw 4 seconds audio fragment and transformed them into the $257 \times 251$ window slices.
Then each time slice of the spectrogram is progressively fed into the encoder module (1 layer LSTM/GRU unit) to a layer of 256 recurrent features $A_h$.
%We compared the performance of LSTM unit and GRU unit in Section \ref{network}.
The height predictor (a 2-layer MLP) takes the recurrent vector as input and performs regression of the temporary length of the air column. The height predictor is supervised with a mean squared error (MSE) loss $\mathcal{L}_{height}$

\begin{equation}
\label{heightloss}
\mathcal{L}_{height} = \|\hat{H}_a - H_a\|^{2}\text{.}
\end{equation}

In addition, leveraging the principle that the liquid height in the target container is monotonically increased, we introduce an auxiliary $\mathcal{L}_{mono}$ to enforce the estimated length of the air column being decreasing along the time $t$
\begin{equation}
\label{monoloss}
\mathcal{L}_{mono} = \sum\limits_t[\max(0, (\hat{H}_{a_{t+1}} - \hat{H}_{a_t}))]\text{.}
\end{equation}

\noindent
\textbf{Overall loss.} Combining with $\mathcal{L}_{height}$ and $\mathcal{L}_{mono}$, the complete training objective for PouringNet is defined by $\mathcal{L}_{audio}$
\begin{equation}
\label{audioloss}
\mathcal{L}_{audio}(\theta)=\mathcal{L}_{height} + \alpha \cdot \mathcal{L}_{mono}\text{,}
\end{equation}
where $\alpha$ is a hyperparameter for balancing these two loss functions. In our implementation, we set it to $0.01$ for the best performances via some preliminary experiments. 

\section{EXPERIMENT}
\subsection{PouringNet Evaluation}
\label{network}
We examined which structure of PouringNet could learn the most indicative representations during the pouring events.
%All proposed models were evaluated on our pouring dataset with the following experiments: 
To explore the appropriate encoder module of the audio branch, we designed a basic feed forward network (two layers of MLP for regression) as one baseline.
And to find out the recurrent units that could maintain the consequent frequency memory better, we evaluated two popular recurrent units: LSTM and GRU.
We refer to these structure as AudioFC, AudioLSTM, and AudioGRU respectively.
There were two evaluation metrics used in this work: 1) the fraction of audio sequence whose the length prediction error is below a threshold; 2) the average length prediction error on each target container (including on all the containers).

\begin{figure*}[t]
	\centering
	\subfigure[]
	{\includegraphics[height=0.22\textwidth]{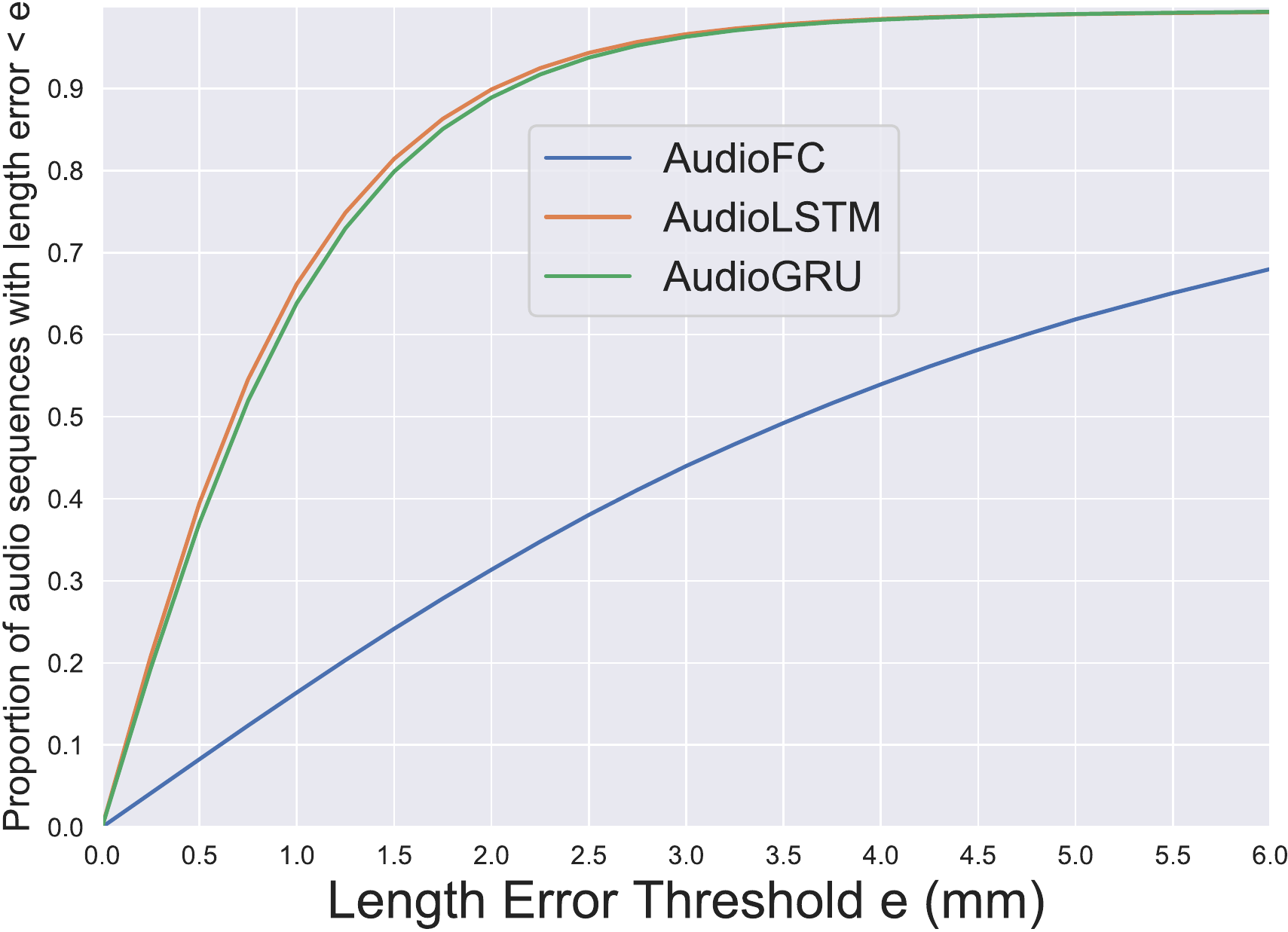}
		\label{fig:acc}}
	\hspace{0.1in}
	\subfigure[]
	{\includegraphics[height=0.22\textwidth]{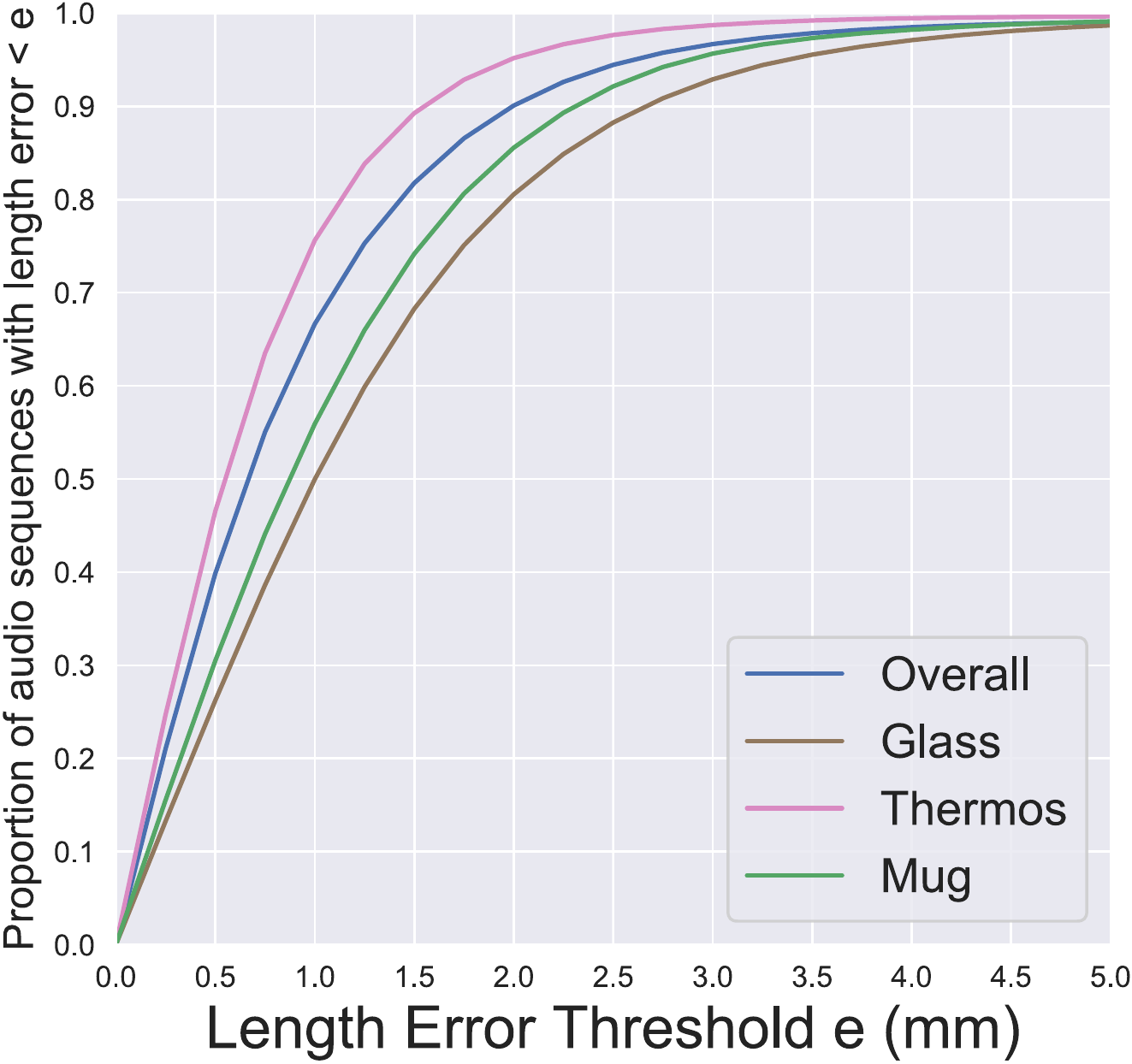}
		\label{fig:bottleacc}}
	\hspace{0.1in}
	\subfigure[]
	{\includegraphics[height=0.22\textwidth]{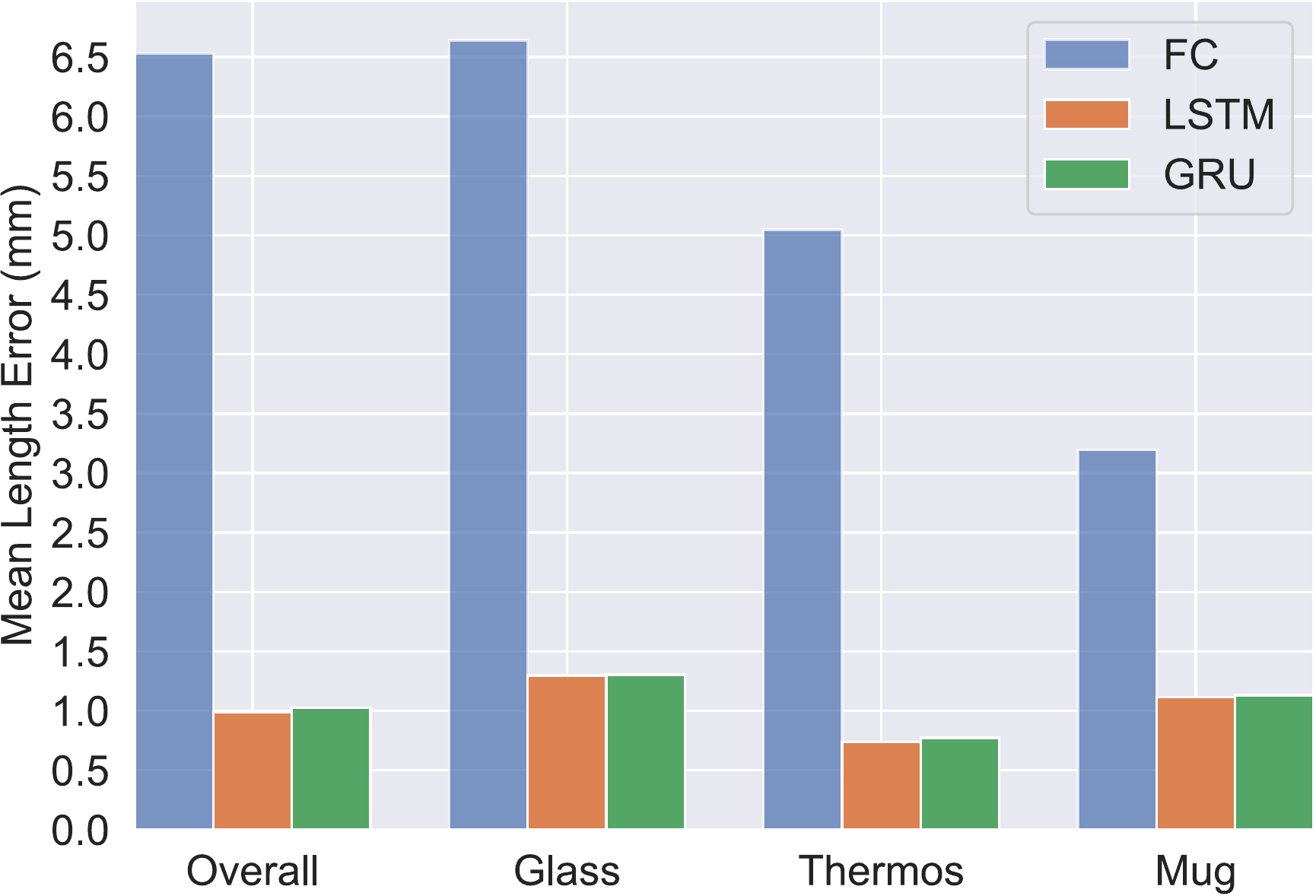}
	\label{fig:errorbar}}
	\caption{Evaluation results of AudioFC, AudioLSTM and AudioGRU models tested on our testing set which collected by human. Evaluation results of the fractions of audio sequence whose the length of the air column's errors are below a threshold \subref{fig:acc} by all models on our overall test data and \subref{fig:bottleacc} only by AudioLSTM model tested on overall test data and each single target container's dataset. \subref{fig:errorbar} Comparison of length errors on the different target containers evaluated by all models.}
	\label{fig:neteval}
	\vskip -0.15in
\end{figure*}

The evaluation results on the testing set are shown in Fig. \ref{fig:acc} and Fig. \ref{fig:errorbar}, which indicate that the recurrent architectures are significantly superior to the AudioFC baseline. 
Both recurrent architectures achieve 90\% accuracy below an absolute 2\,mm length error, and the absolute mean length errors are below 1.5\,mm.
The results not only show that the recurrent architectures can integrate the prior knowledge from audio sequences,
but also verify that the audio vibration is eligible to infer the liquid height between different target containers.
On comparison of these two recurrent architectures, the AudioLSTM architecture slightly outperformed AudioGRU architecture on a single target container and all target containers, demonstrating its advantage of having more trainable parameters when trained with our large training set.

Fig. \ref{fig:bottleacc} manifests the results of AudioLSTM model trained on the training set with only a single target container or all target containers (shown as Overall). As shown in this plot, on the thermos cup it outperforms the other two target containers and the glass shows the worst performance.
This mainly due to the different materials of these containers. For example, the stainless steel material makes the crispest sound.
Furthermore, we can see that the model trained on our complete training set with all types of containers achieves the best accuracy, which suggests that it benefits a lot from a larger dataset and could embody a better-generalized ability than any models that trained on only a part of the dataset.

\subsection{Robotic Experiments}
\begin{figure}[t]
	\centering
		\subfigure[]
		{\includegraphics[height=0.1\textwidth]{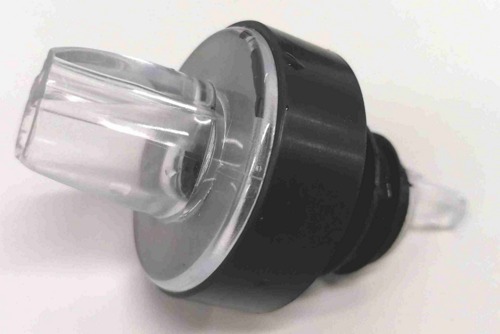}
        \label{fig:spout}}
        %\hspace{0.05in}
        \subfigure[]
        {\includegraphics[height=0.1\textwidth]{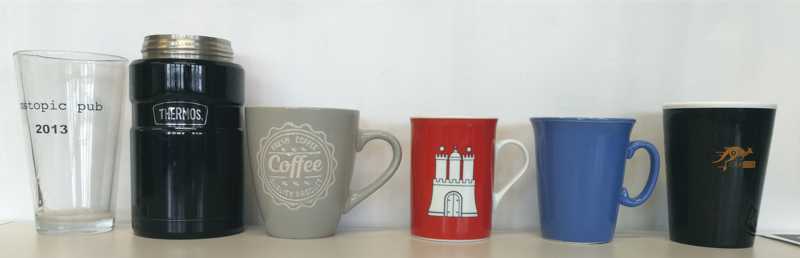}
        \label{fig:newcups}}
    	\caption{\subref{fig:spout} The spout equipped on the source container in robotic experiments. \subref{fig:newcups} From left to right, the first three target containers are the target containers in our datasets: a glass, a stainless steel cup, and a mug; the latter three target containers are the unseen containers used in robotic experiments: a red mug, a blue mug and a plastic cup.
    	}
		\label{fig:robot}
		\vskip -0.15in
\end{figure}

In this section, we evaluate the adaptability and robustness of our audio-based perception method in pouring experiments with a UR5 robot.
In the experiment setup, we fixed the pouring trajectory of the robot and the source container position in the robot’s gripper. 
To make the quality of the recorded pouring sound better, we approximately fixed the center of the gripper right above the top center of the scale at 310\,mm.
Unfortunately, the high position of the gripper gives rise to a high possibility of spillage during pouring.
We solved this problem by equipping the source container used for pouring with a thin spout shown in Fig. \ref{fig:spout}. 
% unknown cups image and s spout image
Accordingly, the pouring speed was drastically reduced due to the spout, compared to the dataset with human pouring. Therefore, we re-collected a small dataset of 30 trials for each target container using the source container with a spout.

%Concerning the slight improvements of multimodal structures and the crucial requirements of the effectiveness in real robot experiments, 
In practice, we first selected the AudioLSTM model which was trained on our original dataset with all target containers to estimate the real-time length of the air column.
Then, we fine-tuned this AudioLSTM model on our new dataset that is collected with a spout on the source container. 
Finally, the refined model was employed as the feedback command, which terminated the pouring immediately once the desired length of the air column had reached.
The average computation time of one feedback loop is 21\,ms, of which nearly 20\,ms are spent on processing the spectrograms (a desktop machine with a 20-core Intel i9-7900X CPU and two 1080Ti GPUs). 

To verify the generalization ability and robustness to robotic pouring tasks, we design four groups of robotic experiments: evaluation on different target containers, different microphone positions, different initial liquid heights and different types of liquid.

\subsubsection{Evaluation of Different Target Containers}
In this experiment, we kept the distance between the target containers and the microphone the same as in our original dataset.
During the robotic pouring, we varied the target length of the air column between [40\,mm, 50\,mm, 60\,mm, 70\,mm, 80\,mm] for three existing target containers in our dataset and three unseen target containers in Fig. \ref{fig:newcups}. The height of the red mug, blue mug and plastic cup respectively are 97\,mm, 94\,mm and 103\,mm.
Owing to the different height of each target container, we also tested 90\,mm and 100\,mm targets for three target containers in our dataset, and a 90\,mm target for the plastic cup.
The water was poured for five times to each considered height of each target container.

\begin{figure*}[t]
	\centering
		\subfigure[]
		{\includegraphics[height=0.25\textwidth]{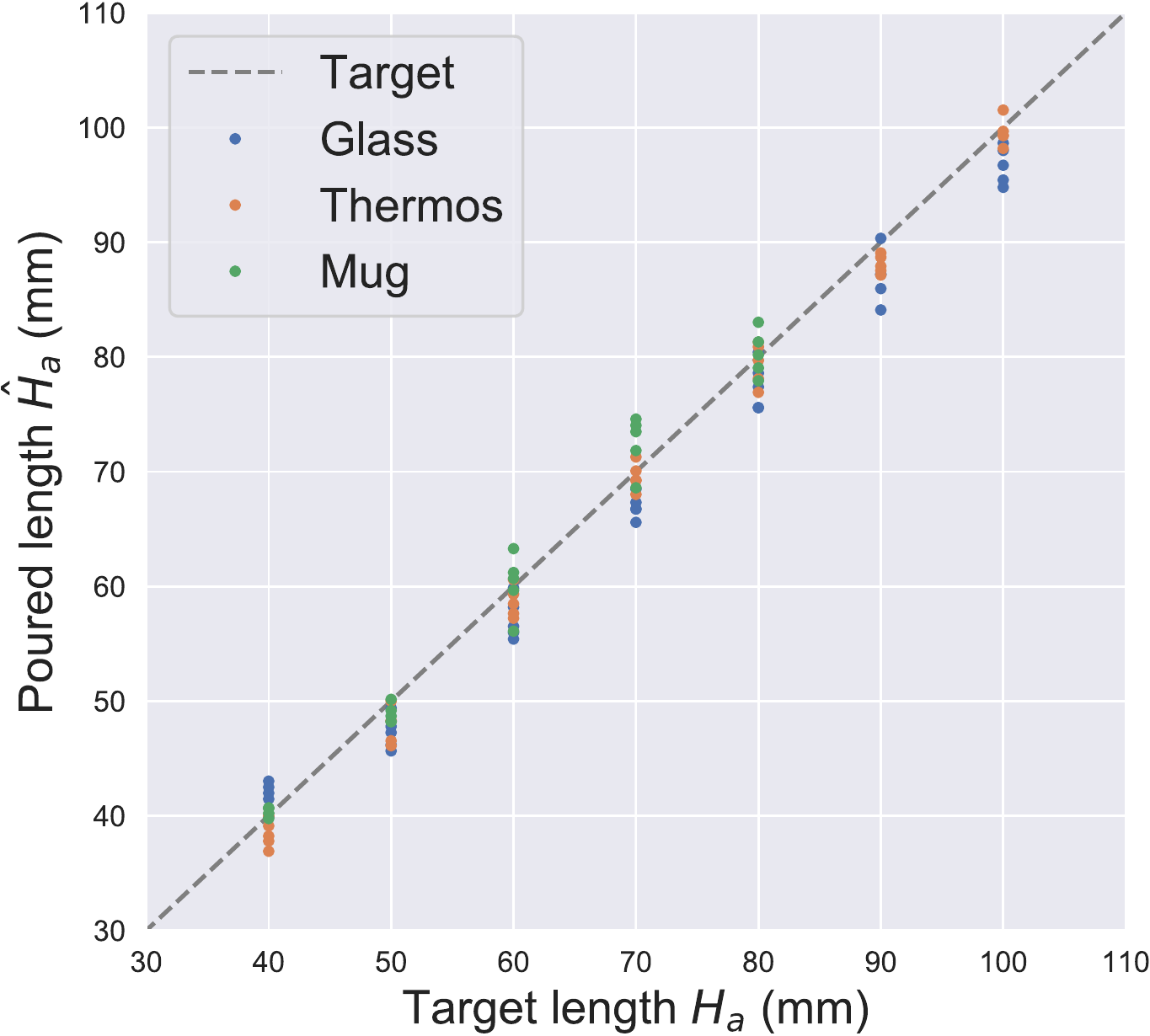}
        \label{fig:lable_cups}}
    %\hspace{0.05in}
    \subfigure[]
    {\includegraphics[height=0.25\textwidth]{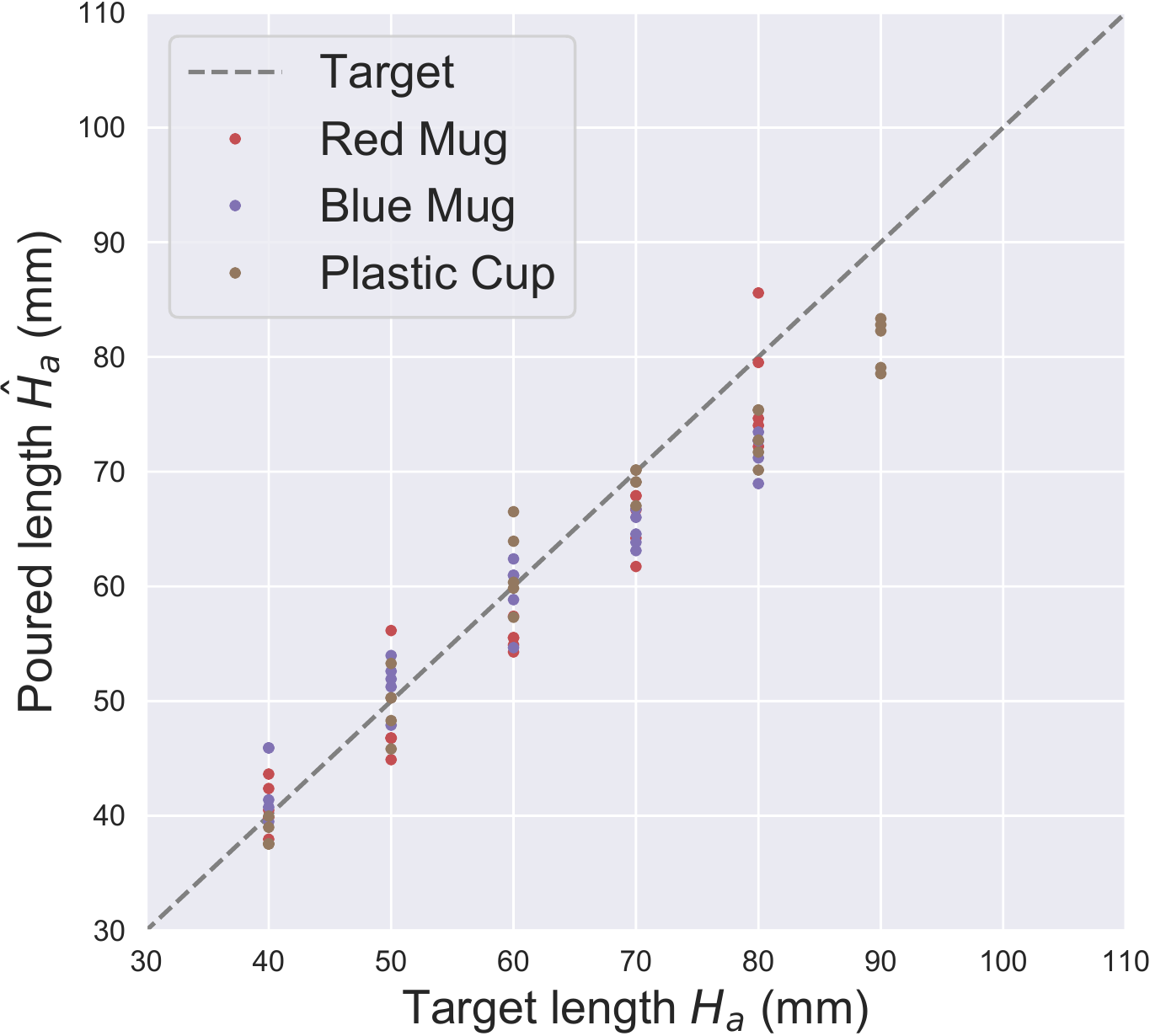}
        \label{fig:unknown_cups}}
	 \subfigure[]
	{\includegraphics[height=0.25\textwidth]{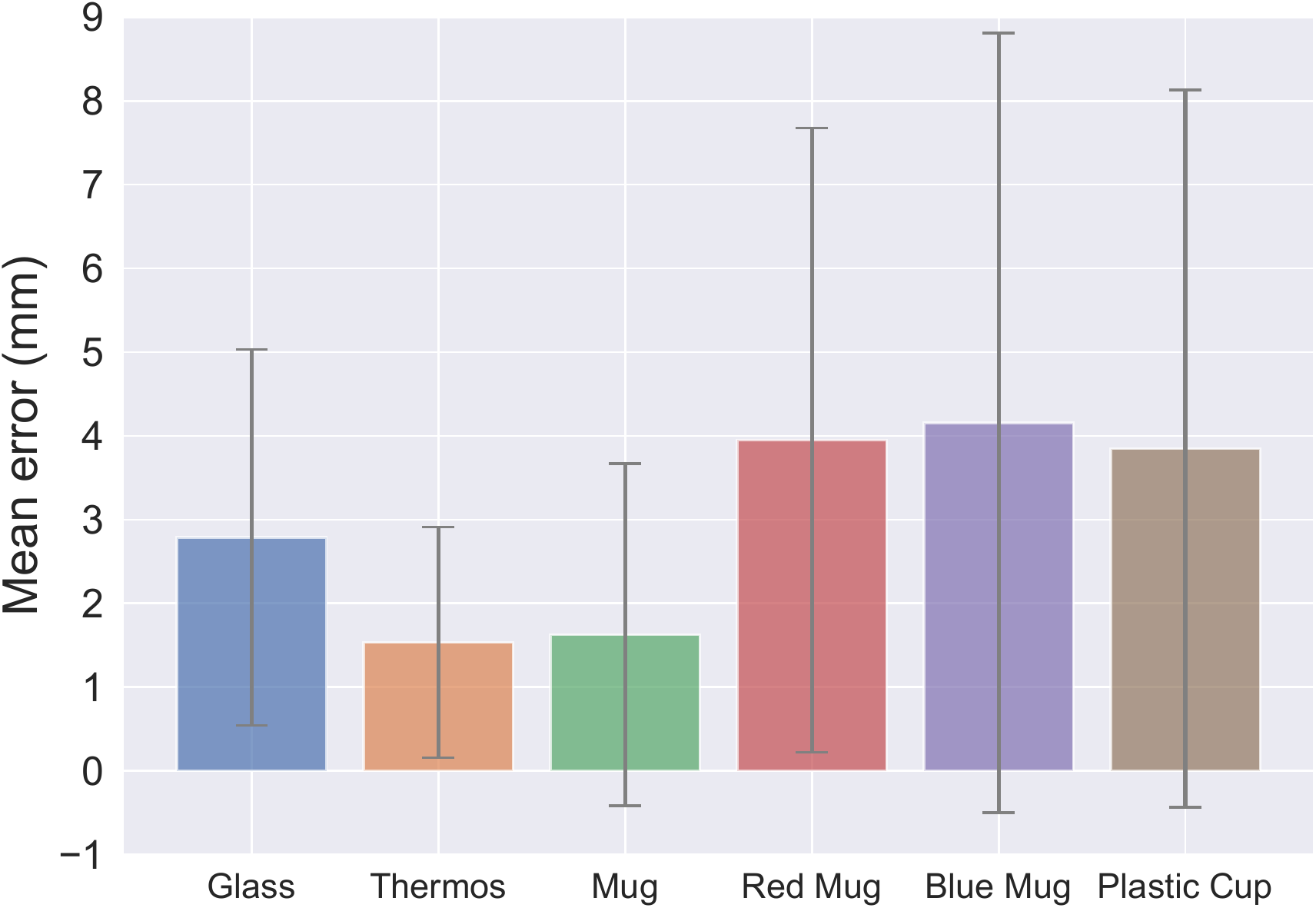}
	%\caption{Comparison on the absolute mean error and standard deviation of the different target bottles.}
	\label{fig:cups_mean}}
	\caption{Controlling results of pouring water to a desired length of air column by trained LSTM model in robotic experiments. \subref{fig:lable_cups} Length estimation of target containers from our dataset. \subref{fig:unknown_cups} Length estimation of unseen target containers. \subref{fig:cups_mean} Comparison on the absolute mean error and standard deviation of different target containers among all desired heights.}
	\label{fig:cups_mean_fig}
	\vskip -0.15in
\end{figure*}

Quantitative results in Fig. \ref{fig:lable_cups} indicate that our audio-based perception system could exploit the useful audio features even though there are huge differences between the human setup (used in our dataset) and the robot setup (used in our robotic evaluations), such as the pouring trajectories and the quiet degree of the pouring environment. From Fig. \ref{fig:lable_cups} and Fig. \ref{fig:unknown_cups}, we can see that when the target length of the air column is shorter (\ie the liquid height is higher), the estimated length of the air column gradually become more accurate. 
% This result is more evident especially on the unseen target containers. 
% Especially for unseen target containers, the bias of the estimated length of the air columns gradually decreases.
It suggests that our model works well in height estimation, and the model will be more sensitive when the target length $H_a$ is relatively high. Since there is more likely to spill out with higher liquid height, this property provides our PouringNet the extra ability to prevent the manipulator from spilling out the liquid.

For numerical results, Fig. \ref{fig:cups_mean} quantifies that the absolute mean errors and the standard deviations of the liquid height are both below 3\,mm among all target containers in human setup and below 4.5\,mm among unseen target containers.
Additionally, we converted the height error of each cup to weight error in this experiment shown in Table \ref{tab:amount}. In previous work on robotic pouring, Schenck \cite{visualpouring1} reported a mean error of 38\,ml and Do \cite{dochau2019} achieved a mean volume error 22.53\,ml over three different target containers. Compared to their results, the robotic pouring with our audio-based perception system can achieve higher precision. 

\begin{table}[ht]
	\centering
	\caption{Absolute Mean Amount Errors and Standard Deviations}
	\vskip -0.15in
	\begin{tabular}{ccc}
	\multicolumn{3}{c}{}\\
	\hlineB{2}
	 Glass & Thermos & Mug \\
	 \hlineB{2}
	$9.54\pm7.81$\,ml & $9.91\pm8.48$\,ml  & $13.79\pm 11.04$\,ml \\
	\hlineB{2}
	\multicolumn{3}{c}{}\\
	 \hlineB{2}
	 Red Mug & Blue Mug & Plastic Cup \\
	\hline
	$7.92\pm7.14$\,ml & $6.42\pm6.31$\,ml  &  $10.72\pm8.70$\,ml
	\\ \hlineB{2}
	\end{tabular}
	\label{tab:amount}
%	\vskip -0.2in
\end{table}

% In summary, PouringNet attains consistent results on robotic evaluation compared with the experiments on dataset, which strongly demonstrate that PouringNet does facilitate a robust and accurate perception for robotic pouring with different target containers.

\subsubsection{Evaluation of Varying Microphone Positions}
\begin{figure}[t]
	\centering
	\subfigure[]
	{\includegraphics[height=0.2\textwidth]{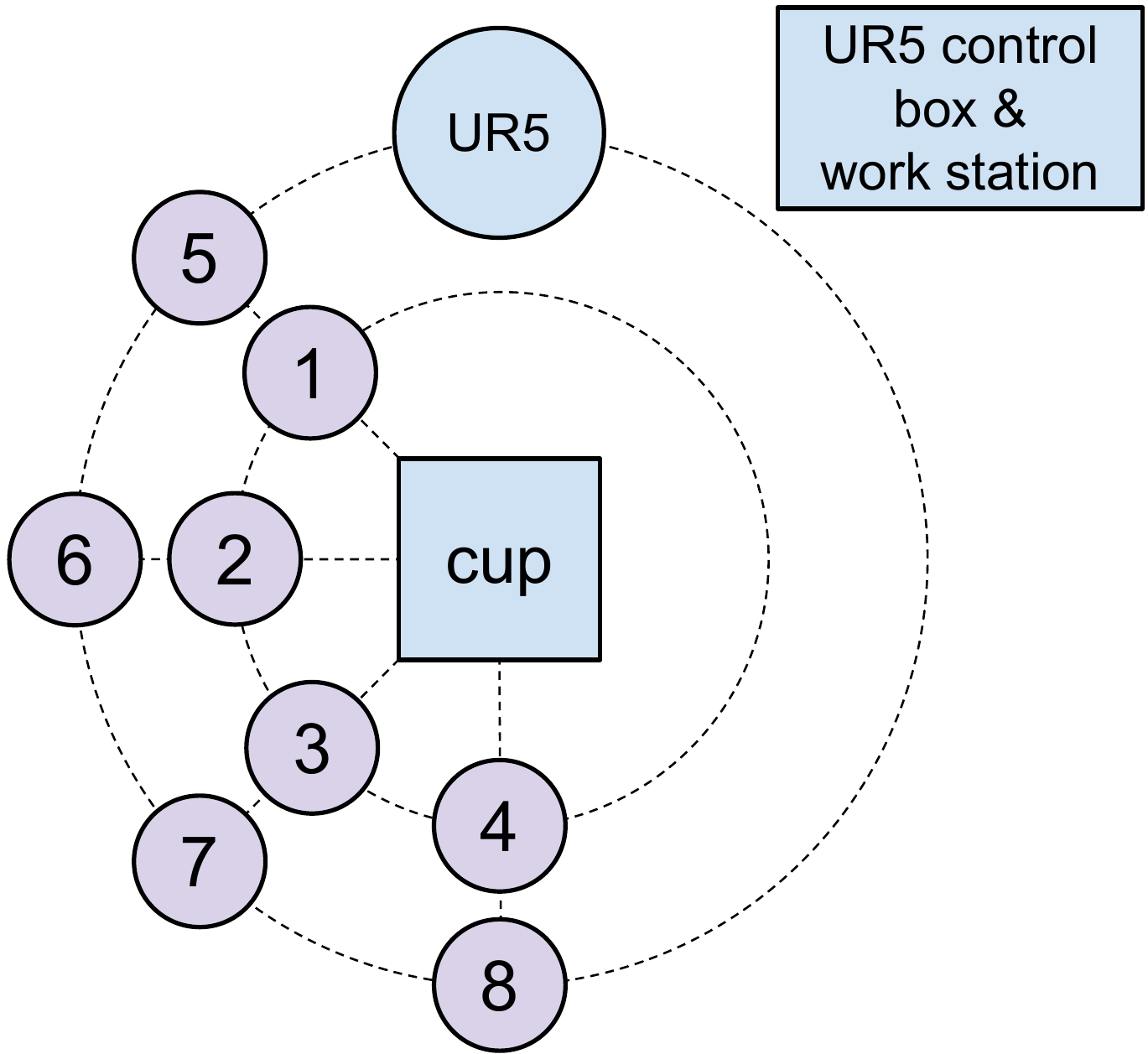}
	\label{fig:pos}}
	%\hspace{0.05in}
	\subfigure[]
	{\includegraphics[height=0.2\textwidth]{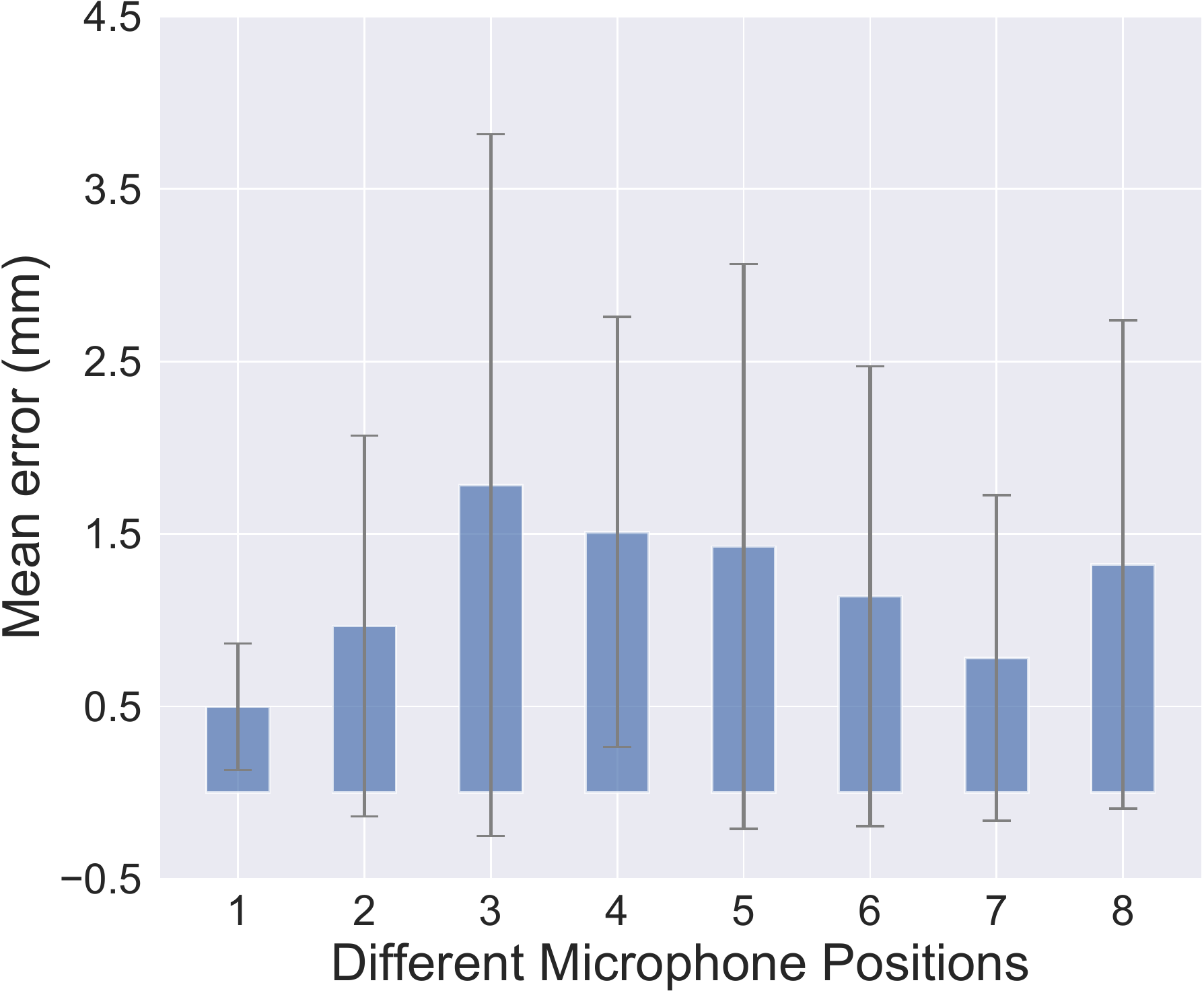}
		\label{fig:pos_eval}}
	\caption{\subref{fig:pos} Schematic diagram of eight microphone positions relative to the target containers, the UR5 robot position, and the control box of UR5 robot (which is the major noise source). \subref{fig:pos_eval} Evaluation results of eight microphone positions shown in \subref{fig:pos}.}
	\vskip -0.15in
\end{figure}

We compared the performance of our method on eight different positions of the bottom center of the microphone as shown in Fig. \ref{fig:pos}. Position 1 is the position that we placed the microphone in our dataset. The distances between the target containers and the bottom center of the microphone at position 1, 2, 3, 4 and at position 5, 6, 7, 8 are 230\,mm and 380\,mm respectively.
We used the mug and the desired length of air column at 40\,mm, as in the later two groups of evaluations. Then we poured water at each microphone position for five times.
Since the robot poured from the right side of the target container, all tested microphone positions were on the left side to avoid a collision between the microphone and the robot.

As shown in Fig. \ref{fig:pos_eval}, the estimation results indicate that our method generalizes well to the different positions of microphone due to the highly consistent of mean height error among all tested positions. In particular, the distance between the noisy robot control box and the microphone does not have a significant influence on the performances, which further suggests the robustness to different quiet degrees of PouringNet.

%Finally, we also study the robustness of our method in the presence of a human voice and soft music to periodically interfere during pouring trajectory roll-out. The policy is able to extract useful sounds from the noisy perturbations. 

\subsubsection{Evaluation of Varying Initial Liquid Height in Target Containers}
\begin{figure}[t]
	\centering
	\subfigure[]
	{\includegraphics[height=0.18\textwidth]{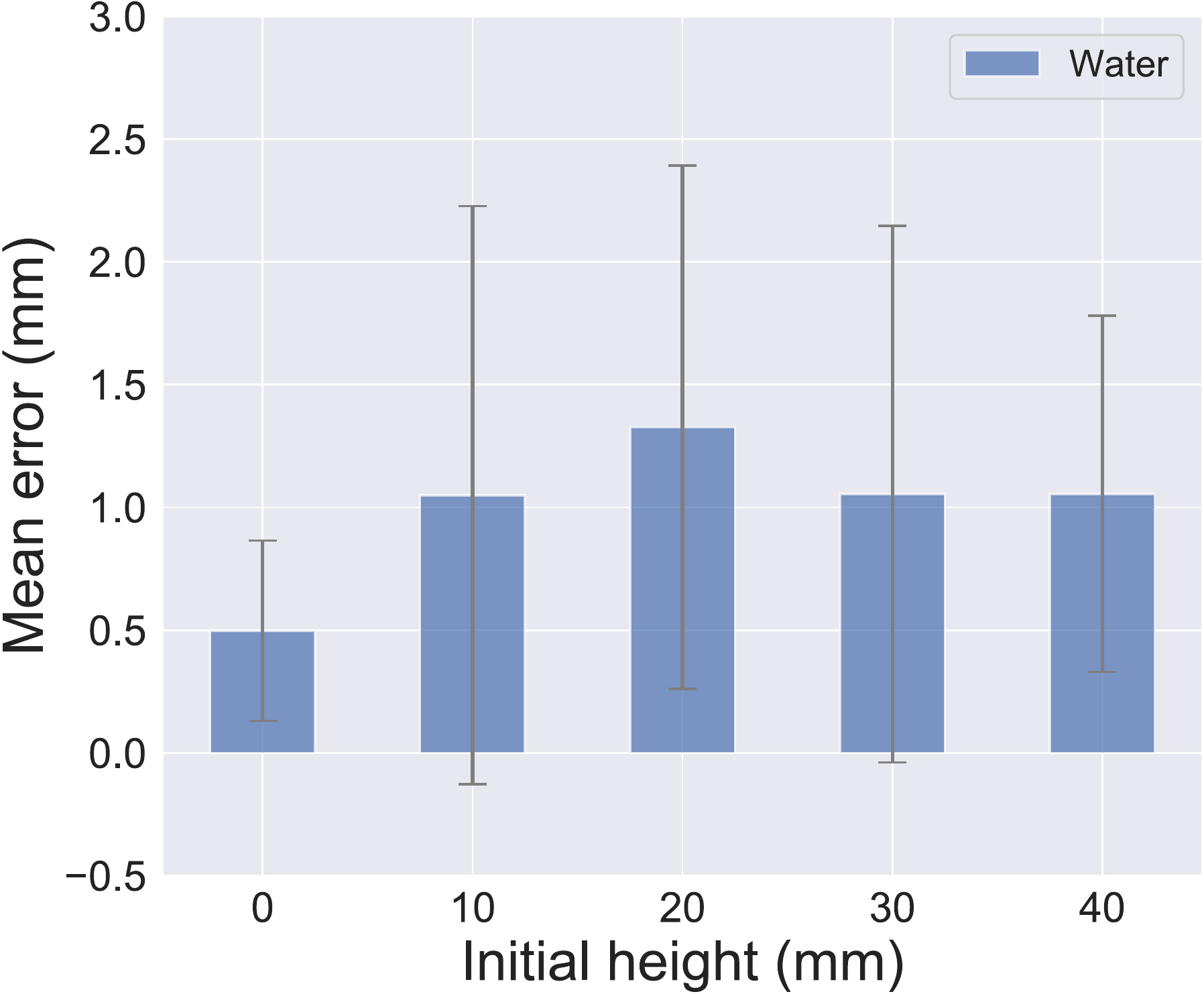}
	\label{fig:inital}}
	\subfigure[]
	{\includegraphics[height=0.18\textwidth]{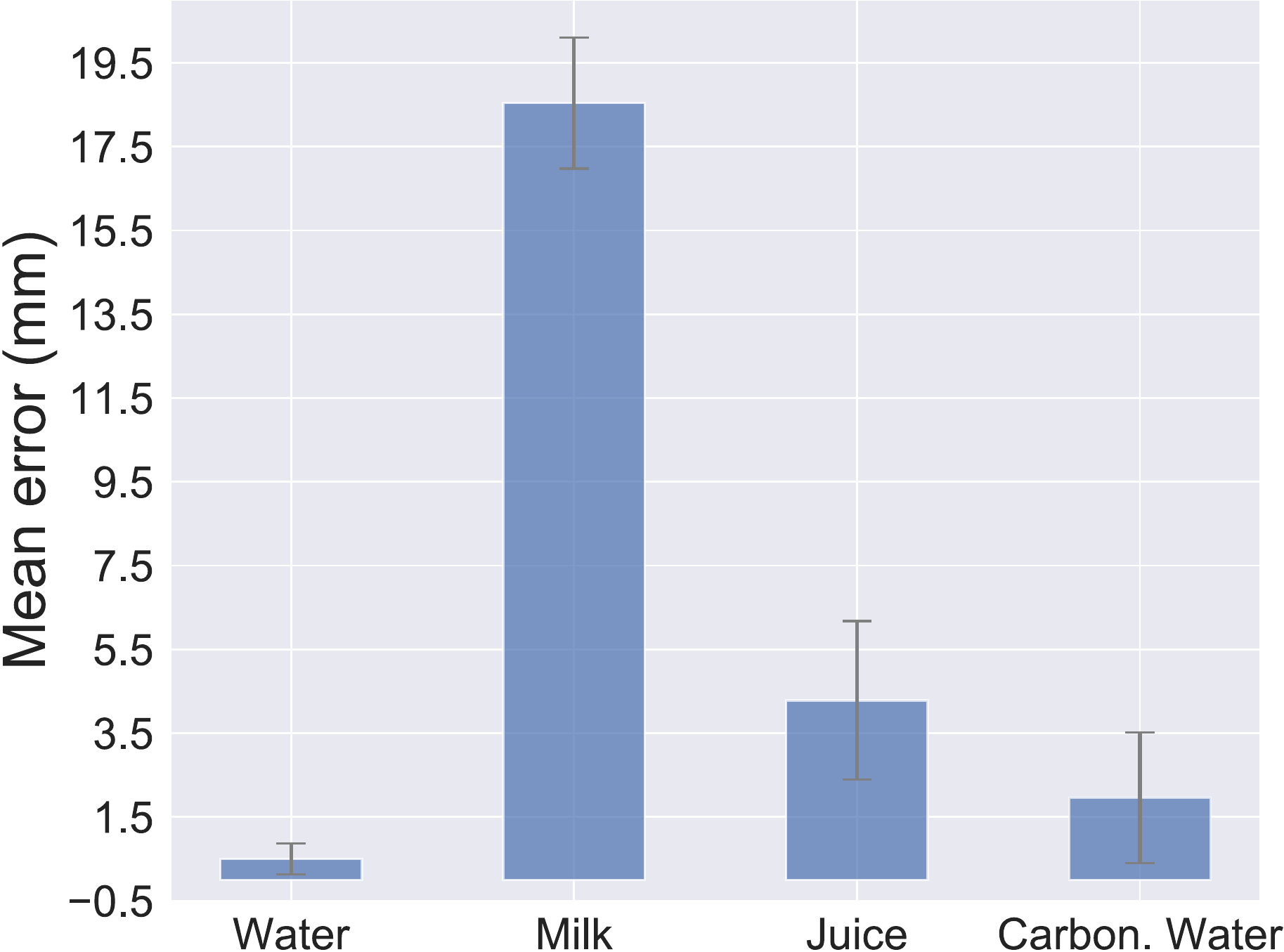}
		\label{fig:diff_liquid}}
	%\hspace{0.05in}
	\caption{Evaluation results of \subref{fig:inital} varying initial height and \subref{fig:diff_liquid} different liquids.}
	\label{fig:nor}
	\vskip -0.15in
\end{figure}

% change a new sentence
In this experiment, we selected the initial liquid height of target containers from the set: $[10\,mm, 20\,mm, 30\,mm, 40\,mm]$, and pouring liquid from each initial length for five times. Other experiment settings are the same as the evaluation of different microphone positions. The results of the absolute mean error and standard deviation are listed in Fig.~\ref{fig:inital}, and it demonstrates that our PouringNet is stable among these different experiment settings, which indicates that it generalizes well to the different initial liquid height in robotic pouring.

\subsubsection{Evaluation of Different Types of Liquid}
\label{samecup}
To analyze the influence of liquid type on our perception method, we also conducted pouring experiments with pure water (which is used in other experiments), carbonated water, 1.8\% fat milk and orange juice with pulp. These types of liquid have different physical properties regarding density, thickness, and viscosity.
We used the same experiment setting as in evaluations of different microphone positions and poured each type of liquid for five times. Fig. \ref{fig:diff_liquid} demonstrates that our model is able to generalize to common household liquids like carbonated water and orange juice.
However, it has failed to work well on milk as its higher viscosity makes the sound weaker and induces difficulties to record and analysis. As the milk and juice are with higher viscosity than pure and carbonated water, it turns out that our the generalization performances of our PouringNet are negatively correlated to the viscosity of the liquid.
%We could imagine that the more viscous the liquid is, the harder our model generalizes.
%We also guess that maybe the temperature difference between milk and the liquids partially causes this poor performance because the temperature of the liquids also represents distinctive audio characteristics \cite{velasco2013sound}.

Through these experiments above on the generalization performances of PouringNet, we %successfully 
verify that our method is able to handle different target containers, microphone positions, initial heights and some liquid types in robotic pouring.
And we also see a restriction of our method is that it cannot apply to liquid with high viscosity. To help to reproduce our results, the related code, dataset, and video are available at \href{https://lianghongzhuo.github.io/AudioPouring}{https://lianghongzhuo.github.io/AudioPouring}.

\section{Conclusion and Future Work} % (0.25 page)
This paper presents a real-time perception system used for estimating the liquid height in robotic pouring.
We offer a multimodal pouring dataset including audio-frequency recordings, liquid real-time weight, force and torque feedback, video and motion trajectories. With this dataset, we develop a robust audio-based perception model named PouringNet.
PouringNet takes audio sequence as input and produces the estimation to the length of the air column of the target containers to represent the height of liquid. 
Especially, using the length of the air column of the target as label could deal with the risk to overfill without any prior knowledge of the target container.
Model evaluations on dataset suggests that the acoustic sequences provide rich information of liquid height and achieve high precision on our test dataset. And various robotic experiments with a focus on generalization and precision performances demonstrate that our proposed PouringNet generalizes well to different experiment settings while the precision can still be guaranteed.
%Compared to other approaches that are either based on visual or haptic sensing, our method can generalize better to different types of containers and different liquids, while the precision of predictions can be guaranteed. 
%In summary, the method can be used for real-time estimation of the liquid height in the robotic environment.

In future work, to facilitate robot behavior in human-robot interaction scenarios, we plan to extend our approach to more noisy environments with human voice and make the distance between the source and target container changeable.
%Classify liquid type from the audio is also possible.
In addition, making use of the force, motion trajectories and visual data from our multimodal dataset and studying the complementarity and interaction between multiple modalities in robotic pouring would also be an exciting direction of future research. 

\small{
\section*{ACKNOWLEDGMENT}
This research was funded by the German Research Foundation (DFG)
and the National Science Foundation of China (NSFC)
in project Crossmodal Learning, DFG TRR-169/NSFC 61621136008. 
And it was partially supported by project STEP2DYNA (691154).
}

\bibliographystyle{IEEEtran} % (0.5 page, about 20 refs)
\bibliography{IEEEabrv,ref}

\end{document}